\documentclass[10pt,twocolumn,letterpaper]{article}

\usepackage[pagenumbers]{cvpr} 

\usepackage{graphicx}
\usepackage{amsmath}
\usepackage{amssymb}
\usepackage{booktabs}
\usepackage{multirow}

\usepackage[pagebackref,breaklinks,colorlinks]{hyperref}

\usepackage[capitalize]{cleveref}
\crefname{section}{Sec.}{Secs.}
\Crefname{section}{Section}{Sections}
\Crefname{table}{Table}{Tables}
\crefname{table}{Tab.}{Tabs.}

\def\cvprPaperID{10429} 
\def\confName{CVPR}
\def\confYear{2023}

\begin{document}

\title{Decomposed Soft Prompt Guided Fusion Enhancing for Compositional \\ Zero-Shot Learning}


\author{Xiaocheng Lu\footnote[2]{}~\textsuperscript{1}, Ziming Liu\footnote[2]{}~\textsuperscript{2}, Song Guo\textsuperscript{2}, \textit{Fellow, IEEE}, Jingcai Guo\textsuperscript{2}, \textit{Member, IEEE}\\
\textsuperscript{1}School of Artificial Intelligence, Optics and Electronics, Northwestern Polytechnical University\\
\textsuperscript{1,2}Department of Computing, The Hong Kong Polytechnic University\\
xiaochenglu1997@gmail.com, \{ziming.liu, jingcai.guo\}@connect.polyu.hk, song.guo@polyu.edu.hk
}

\maketitle

\begin{abstract}
Compositional Zero-Shot Learning (CZSL) aims to recognize novel concepts 
formed by
known states and objects during training.
%
Existing methods either learn the combined state-object representation, challenging the generalization of unseen compositions, 
or design two classifiers to identify state and object separately from image features, ignoring the intrinsic relationship between them. 
To jointly eliminate the above issues and construct a more robust CZSL system, we propose a novel framework termed \underline{\textbf{D}}ecomposed \underline{\textbf{F}}usion with \underline{\textbf{S}}oft \underline{\textbf{P}}rompt (DFSP)\footnote{Code is available at: \url{https://github.com/Forest-art/DFSP.git}}, by involving vision-language models (VLMs) for unseen composition recognition. 
Specifically, DFSP constructs a vector combination of learnable soft prompts with state and object to establish the joint representation of them. 
In addition, a cross-modal decomposed fusion module is designed between the language and image branches, which decomposes state and object among language features instead of image features. 
Notably, being fused with the decomposed features, the image features can be more expressive for learning the relationship with states and objects, respectively, to improve the response of unseen compositions in the pair space, hence narrowing the domain gap between seen and unseen sets.
%
%
Experimental results on three challenging benchmarks demonstrate that our approach significantly outperforms other state-of-the-art methods by large margins. 
\end{abstract}

\begin{figure}[t]
  \centering
  \includegraphics[scale=0.33]{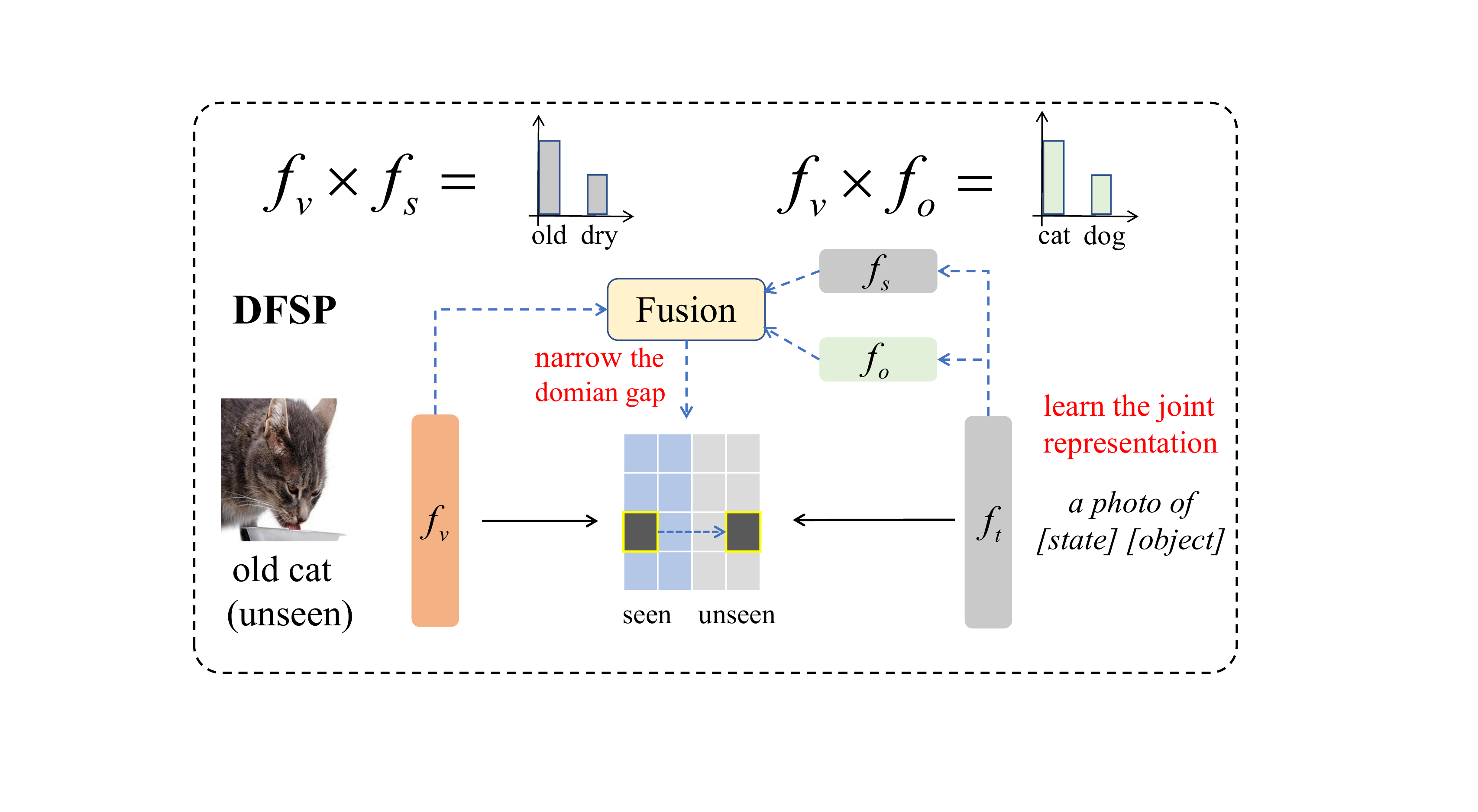}
   \caption{The overview of DFSP. Our method aims to narrow the domain gap between seen and unseen compositions by fusing decomposed features $f_o$ and $f_s$ with image feature $f_v$, while learn the joint representation between state and object in language branch. Being fused with the state and object features, image feature can learn the response of them respectively and improve the sensitiveness of unseen compositions.}
   \label{fig:overview}
\end{figure}

\section{Introduction}
\label{sec:intro}

Given an unseen concept, such as \textit{green tiger}, even though this is a nonexistent stuff humans have never seen, they may associate the known state \textit{green} with an image of \textit{tiger} immediately.
Inspired by this, Compositional Zero-Shot Learning (CZSL) is proposed with the purpose of equipping models with the ability to recognize novel concepts generated as humans do. Specifically, CZSL learns on visible primitive composed concepts (state and object) in the training phase, and recognizes unseen compositions in the inference phase.
Due to the combination of state and object, CZSL aims to learn the joint-representation of them and improve the generalization from seen to unseen compositions.


Some prior algorithms~\cite{li2020symmetry, misra2017red} design two classifiers to identify state and object separately, while these models overlook the intrinsic relation between them. After the primitive concepts are obtained, the association between state and object could be established again through graph neural network (GNN)~\cite{mancini2022learning} or external knowledge compositions~\cite{karthik2022kg}. Nevertheless, these are post-processing methods and these classifiers are separated from image features with strong correlation, ignoring entanglement.
Some other methods~\cite{nagarajan2018attributes, nan2019recognizing} are to directly treat the combination as an entity, converting CZSL into a general zero-shot recognition problem.
Generally, the visual features are projected into a shared semantic space and the distance between entities is optimized, such as Euclidean distance~\cite{yang2022decomposable}.
If too much attention is paid to the composed concepts in the training stage, the model can not be generalized well to unseen compositions, causing the domain gap between seen and unseen sets.
In summary, these methods are all visual recognition models, which are limited by the strong entanglement of states and objects in image features.

\begin{figure*}[htp]
  \centering
  \includegraphics[scale=0.53]{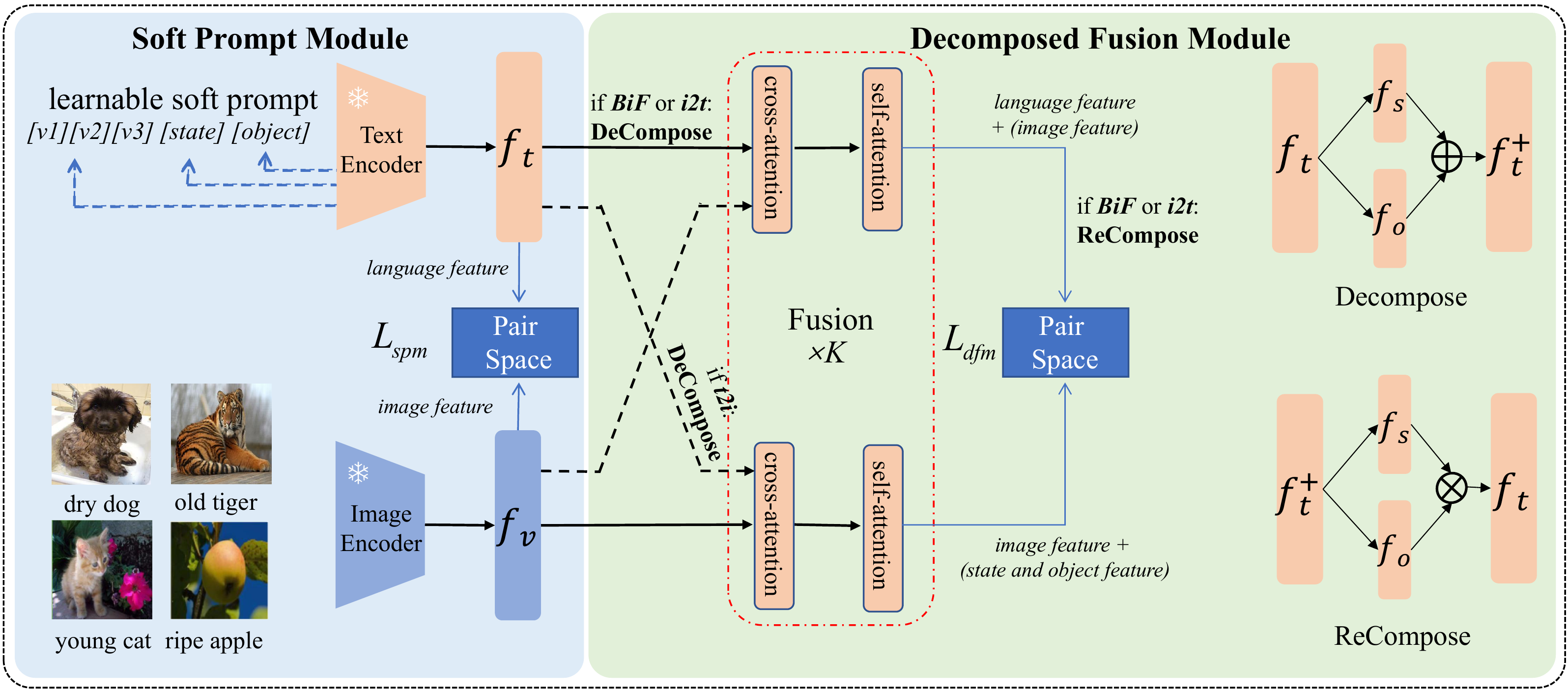}
   \caption{The framework of our proposed DFSP, which consists of Soft Prompt Module (SPM) and Decomposed Fusion Module (DFM). Since DFSP is vision-language model, it could also be divided into two branches, language segment and image segment. SPM aims to construct and preserve the joint representation of state and object, then convert the discrete prompt to learnable soft prompt, causing the extracted language features $f_t$ more discriminative and more suitable for new tasks, especially CZSL. There are three forms of decomposed fusion in SPM, \textit{BiF}, \textit{i2t} and \textit{t2i} respectively, and if the fusion method is \textit{t2i}, only decomposition exists on the language to image branch.
   Meanwhile, decomposition and recomposition coexist in the fusion method of \textit{BiF} and \textit{i2t}.
   After decomposing language feature into independent state feature $f_s$ and object feature $f_o$, DFM fuses them with image feature $f_v$ and calculates similarity in the final pair space. DFSP can not only learn the joint representation of state and object, but also shrink the domain gap of seen and unseen composition sets.}
   \label{fig:framework}
\end{figure*}

In contrast, we focus on designing novel approaches based on vision-language models (VLMs) to cope with CZSL challenges. Since state and object are two separate words in the text, they are less entangled in language features than image features and could be decomposed more easily and precisely. 
Certainly, state and object are also intrinsically linked in the text, such as \textit{ripe apple} instead of \textit{old apple}. Constructing the combination in the form of text can also establish the joint representation of state and object to pair with images. 
Meanwhile, the decomposed state and object features can also be independently associated with the image feature, easing the excessive bias of the model towards seen compositions and enhancing the unseen response (shown in Fig. \ref{fig:overview}).
To improve CZSL with VLMs, we design \textbf{D}ecomposed \textbf{F}usion with \textbf{S}oft \textbf{P}rompt (DFSP), an efficient framework aimed to both learn about the joint representation of primitive concepts and shrink the domain gap between seen and unseen composition sets, as shown in Fig.~\ref{fig:framework}.
To be specific, DFSP is designed as a fully learnable soft prompt including prefix, state and object, which constructs the joint representation between primitive concepts and can be fine-tuned well for new supervised tasks.
We then design a decomposed fusion module (DFM) for state and object, which decomposes features extracted from text encoder, such as Bert~\cite{devlin2018bert}, etc.
Meanwhile, the decomposed language features and image features of DFSP interact with information in a cross-modal fusion module, which is crucial for learning high-quality language-aware visual representations.
During the phase of fusion, the image can establish separate relationships with the state and object, and then is paired with the composed prompt feature in the pair space, improving its response even for unseen compositions to shrink the domain gap.

Generally, this paper makes the following contributions:

\begin{itemize}
	\item A novel framework named Decomposed Fusion with Soft Prompt (DFSP) is proposed, which is based on vision-language paradigm aiming to cope with CZSL.
    \item The Decomposed Fusion Module is designed for CZSL specifically, which decomposes the concepts of language features and fuses them with image features to improve the response of unseen compositions.
	\item We design a learnable soft prompt to construct the joint-representation of state and object, which can be more precisely decomposed than images.
    \item Extensive experiments demonstrate the effectiveness of DFSP, which greatly outperforms the state-of-the-art CZSL approaches on both closed-world and open-world.
\end{itemize}

\section{Related Work}
We describe the compositional zero-shot learning and prompt learning in this section.

\textbf{Compositional Zero-Shot Learning.} 
CZSL~\cite{gu2021class, li2020symmetry, mikolov2013distributed, misra2017red, nagarajan2018attributes} is a task similar to how humans can imagine and discriminate unseen concepts according to the concepts they have learned, which is a significant branch of ZSL~\cite{guo2023graph, lampert2013attribute, li2021generalized, wei2020lifelong, chen2022duet, liu2021goal, guo2020novel, 9953574}. For more details, typical ZSL utilizes attributed vectors to recognize unseen objects, while the classes in CZSL are described as state-object pairs.

For CZSL, early works learn a classifier for recognition and a transformation module to convert state or object~\cite{misra2017red, nagarajan2018attributes}. 
Some recent works utilize two separate classifiers to recognize state and object respectively~\cite{li2020symmetry, misra2017red, li2022siamese, karthik2022kg}.
Also, some works combine the encoded attribute/state and object features with late fusion by using multi-layer perceptron~\cite{purushwalkam2019task}.
Li \textit{et al.} introduce contrastive learning into CZSL, and design a siamese network to identify state and object in the contrastive space, respectively~\cite{li2022siamese}.
Other methods~\cite{nagarajan2018attributes, nan2019recognizing} focus on the joint representation of the compositions, which 
learn an emdedding space to map the compositions like ZSL.
Recent works utilize graph networks to represent the state and object relationship and then learn their compositions~\cite{mancini2022learning, ruis2021independent, wu2019simplifying}.
Besides, Nihal \textit{et al.} first attempt to use a VLM model for CZSL, replacing the classes in prompt with a learnable combined state and object vector representation~\cite{nayak2022learning}.

Also, for the possibility of compositions during testing, it is sometimes not limited to compositions within the testing data, but tend to the real world where all compositions should be considered. As opposed to the previous closed-world, some work aim at open-world by using external knowledge to filter infeasible compositions~\cite{mancini2021open, mancini2022learning, karthik2022kg}.

\textbf{Prompt Learning.}
Prompt Learning refers to processing the input text through a specific template, and reconstructing the task into a form that can more fully utilize the pre-trained language model~\cite{bach2022promptsource, brown2020language, bommasani2021opportunities, sanh2021multitask, vu2021spot, zhou2022learning}. 
Prompting makes the pre-training model and downstream tasks closer, which is different from fine-tuning. 
Fine-tuning uses pre-training models for downstream tasks, while prompt learning is to adapt various downstream tasks to pre-trained models by reconstructing them.
Benefiting from having pre-trained on a large-scale data and associating with multi-modal information, prompt learning can achieve great performance in zero-shot and few-shot on a wide range of tasks~\cite{qin2021learning, radford2021learning}.

Take the CLIP~\cite{radford2021learning} model as an example, discrete prompt has difficulty performing well on downstream tasks even when trained on new data. Soft prompt is another paradigm of prompt learning, where the part of the prompt can be fine-tuned on new challenging tasks~\cite{lester2021power, liu2021pre}.
Some recent works utilize soft prompt to improve downstream tasks and reach fine performance~\cite{lester2021power, li2021prefix, shin2020autoprompt}.
CoOp~\cite{zhou2022learning} convert the prefix part of the prompt to soft prompt like \textit{[v1][v2][v3]object}, fix the parameters of other parts, and only fine-tune the prompt.
In contrast, CSP~\cite{nayak2022learning} sets the primitive concepts section of the prompt to be soft like \textit{a photo of [state][object]}.
While these methods focus only on a certain part of the prompt, we soften the entire prompt to better fine-tune in the new scenario.

\section{Approach}
For CZSL, entities exist in the form of a combination of state and object, and the model needs to be trained on the seen composition set while tested on the unseen set.
To address this challenge, we propose a novel formulation termed Decomposed Fusion with Soft Prompt (DFSP), which constructs a vision-language paradigm with soft prompt and decomposed fusion module.
DFSP first construct soft prompt with state and object to establish the joint representation of them.
Meanwhile, DFSP decomposes language features to separated state and object features, and utilizes cross-modal fusion to  transfer knowledge between decomposed language and images.
The framework of our proposed method is shown in Fig. \ref{fig:framework}.

\subsection{Problem Formulation}
Given state set $\mathcal{A} = \left \{ s_{0},s_{1},\dots ,s_{n}  \right \} $ and object set $\mathcal{O} = \left \{ o_{0},o_{1},\dots ,o_{m}  \right \} $ as the primitive concepts of CZSL, we can compose them as a composition set $\mathcal{C} = \mathcal{A}  \times \mathcal{O}$, where the size of $\mathcal{C}$ is $n \times m$.
Besides, we denote two disjoint sets $\mathcal{C}^{s}$ and $\mathcal{C}^{u}$, where $\mathcal{C}^{s}$, $\mathcal{C}^{u}$ are subsets of the composition set $\mathcal{C}$ and $\mathcal{C }^{s} \cap \mathcal{C}^{u} = \phi  $.
Specifically, $\mathcal{C}^{s}$, $\mathcal{C}^{u}$ represent the seen and unseen sets, respectively, where $\mathcal{C}^{s}$ is used for training and $\mathcal{C}^{u}$ is used for testing.
$\mathcal{T} = \left \{ \left ( x_{i},c_{i} | x\subset \mathcal{X}, c\subset \mathcal{C}^{s}      \right )  \right \} $ is the training set where $\mathcal{X}$ is the input image space and $c$ belongs to the seen composition label set.

The CZSL task aims to train a model $\mathcal{M}:\mathcal{X} \to \mathcal{C} ^{t}$ to predict compositions in the test samples space $\mathcal{C} ^{t}$. 
If $\mathcal{C} ^{t} \cap \mathcal{C} ^{s} \equiv \phi $, where the model only predicts unseen compositions.
Follow the setting of Generalized ZSL~\cite{xian2018zero}, testing samples contain seen and unseen compositions, i.e., $\mathcal{C} ^{s} \cup \mathcal{C}^{u} $ in this paper.
Generally, when testing, only the known composition space of test samples is required, which is called closed-world. 
For Open-World~\cite{karthik2022kg}, the composition space for testing is all possible combinations, i.e., $\mathcal{C}^{t} = \mathcal{C}$.

\subsection{Decomposed Fusion with Soft Prompt Network}
Given an image such as \textit{dry dog}, \textit{dry} and \textit{dog} have strong joint representation in image features, which is entanglement of state and object.
Directly identifying the state and object of an image feature separately will lose its joint representation.
For a short sentence of \textit{dry dog}, although its embedding is also a combination of \textit{dry} and \textit{dog}, this combination is not strong entangled and could be precisely decomposed.
Inspired by this, we propose a vision-language paradigm for CZSL, called Decomposed Fusion with Soft Prompt (DFSP), which includes two modules, Soft Prompt Module (SPM) and Decomposed Fusion Module (DFM). SPM is responsible for the construction of joint representation between state and object, and DFM is to decompose and fuse language features with image features to improve the sensitiveness of unseen compositions.
The overall architecture of DFSP is shown in Fig. \ref{fig:framework}.

\textbf{Soft Prompt Module.}
DFSP is a vision-language model, in which encoders utilize Contrastive Language-Image Pre-Training (CLIP)~\cite{radford2021learning}, which has pretrained on nearly 400M text-image pairs.
The feature extracted section of DFSP consists of the image encoder and the text encoder.
For the image encoder, it can be a vision transformer (ViT)~\cite{dosovitskiy2020image} or a convolutional neural network~\cite{he2016deep}, while the text encoder includes several transformer encoder layers.
It is worth mentioning that the parameters of image encoder and text encoder are frozen, because the CLIP model has done enough pre-training, we only need to fine-tune for downstream tasks.

To be specific, the entities consist of state and object in CZSL are transformed into natural language prompts like \underline{\textit{a photo of [state][object]}}, compared with \underline{\textit{a photo of [class]}} in CLIP.
Before extracting the text representations,
the prompts need to be converted to tokens for each word by tokenizer and the embedding function maps the tokens to the vocabulary.
Due to the discrete vocabulary \textit{[state][object]}, the model cannot be adapted to CZSL well.
CSP~\cite{nayak2022learning} compares the CLIP model and soft \textit{[state][object]} in prompt, which achieves some progress.
Nevertheless, while CSP works well for CZSL, prefix is still fixed, which is too dependent on state and object learning for the model. And a prefix like \textit{a photo of} is not necessarily the best prompt form; if it can also be updated, the model can be generalized better.

In DFSP, the prompt is fully learnable soft prompt \underline{\textit{[v1][v2][v3][state][object]}}.
First, we build a prompt set with a prefix context, state and object, which is formulated as follows:
\begin{equation}
    P(s,o) = \left \{ x_{0},x_{1},\dots,x_{p},x_{s},x_{o}   \right \} ,
    \label{equ: prompt}
\end{equation}
where $\left \{ x_{0}, \dots, x_{p} \right \}$ is the prefix context and the $x_{s}$ and $x_{o}$ represents the state and object vocabulary for the composition set $P(s, o)$.
Then, the prompt will be converted to learnable embeddings as follows:
\begin{equation}
    P^{soft}  = \Gamma  (P(s,o)) = \left \{ \theta _{0},\theta _{1},\dots,\theta _{p},\theta _{s},\theta _{o}   \right \} , 
\end{equation}
where $\Gamma$ is the embedding function, $\left \{ \theta_{0}, \dots,\theta_{p} \right \}$ is the learnable prefix context and $\theta_{s}$ and $\theta_{o}$ denotes the learnable state and object embeddings.
For the training samples $x\in \mathcal{X}$ and the soft prompt $P^{soft}$, we can extract the image and language features:
\begin{equation}
    f_{v} =\frac{E_{v} (x)}{\left \|  E_{v} (x) \right \| }, f_{t} = \frac{E_{t} (P^{soft})}{\left \|  E_{t} (P^{soft}) \right \| },
\end{equation}
where $E_{v}$ and $E_{t}$ represent the image encoder and text encoder, and $\left\| \cdot  \right\|$ means the norm calculation. Next, we can compute the class probability $p_{spm}(\frac{y={(s,o)}}{x:\theta } )$ as follows:
\begin{equation}
    p_{spm}(\frac{y={(s,o)}}{x;\theta } )=\frac{exp(f_{v}\cdot f_{t})}{\sum _{(\bar{s},\bar{o} ) \in \mathcal{C}^{s} }exp(f_{v} \cdot  f_{t}) } .
\end{equation}

Finally, we can minimize the cross entropy loss in the soft prompt module:
\begin{equation}
    \mathcal{L}_{spm}= -\frac{1}{\left | \mathcal{C}^{s} \right | }\sum_{(x,y)\in \mathcal{C}^{s} }log \left ( p_{spm} (\frac{y={(s,o)}}{x;\theta } )\right )    .
\end{equation}

\textbf{Decomposed Fusion Module.}
The language and image representations of SPM are directly calculated their similarity in the pair space, causing the model tend to the seen compositions of the training set, lacking the ability to perceive unseen samples.
To narrow the domain gap between seen and unseen sets, we propose Decomposed Fusion Module (DFM), which decomposes language features into state and object features and fuses them with image feature.

The design of DFM has three key points for DFSP: 
i) DFM only decomposes the language features, in which state and object has less entanglement. Since state and object are separated in text, they can be decomposed easily and precisely, which preserves the joint representation of state and object; 
ii) The decomposed state and object features are fused with image features to realize information interaction between the two modalities. Meanwhile, DFM establishes respective associations of the image with the state and object, improving the responsiveness in the pair space, especially for the unseen compositions; 
iii) If there are many categories of states and objects, the composition set will be particularly large, limiting the performance of the model. DFM can reduce the complexity of the composition, from $O(n \times m)$ to $O(n + m)$. Since training is on the seen set and testing is on the unseen set, the number of compositions is inconsistent. For cross-modal fusion, decomposition is essential to ensure that the model can maintain a fixed amount of parameters when the composition set changes.

With SPM, we can extract the image feature $f_{v}$ and language feature $f_{t}$, and then the state feature $f_{s}$ and object feature $f_{o}$ can be decomposed as follows:
\begin{equation}
    f_{s}, f_{o} = \mathcal{D}_{e}(f_{t}),
\end{equation}
where $De(\cdot)$ denotes the decomposition, and it's formulation is:
\begin{equation}
    \mathcal{D}_{e} = \left \{ \sum_{i}\frac{f_{t}}{\left | j \right | },
\sum_{j}\frac{f_{t}}{\left | i \right | }  | i \in \mathcal{S}, j \in \mathcal{O}, (i,j) \in \mathcal{C}^{s}  \right \}.
\end{equation}
The language feature $f_{t}$ is the combined seen feature set, and $\mathcal{D}_{e}$ calculates its average state feature relative to each object and the average object feature of each state.
The class scale of them is $n$ and $m$ respectively, and then $f_{s}$, $f_{o}$ will be concatenated to $f^{+}_{t}$, which is consistent during training and testing.

The decomposed state and object features can also be supervised to provide some guidance for subsequent training.
We can compute the state probability  $p(\frac{y={s}}{x:\theta } )$ and object probability $p(\frac{y={o}}{x:\theta } )$ as follows:
\begin{equation}
    p(\frac{y={s}}{x;\theta } )=\frac{exp(f_{v}\cdot f_{s})}{\sum _{(\bar{s}) \in \mathcal{A} }exp(f_{v} \cdot  f_{t}) } ,
\end{equation}
\begin{equation}
    p(\frac{y={o}}{x;\theta } )=\frac{exp(f_{v}\cdot f_{o})}{\sum _{(\bar{o} ) \in \mathcal{O} }exp(f_{v} \cdot  f_{o}) } .
\end{equation}
And the cross entropy loss can be minimized by:
\begin{equation}
\begin{split}
    \mathcal{L}_{st+obj}= -\frac{1}{\left | \mathcal{A} \right | }\sum_{(x,y)\in \mathcal{C}^{s} }log \left ( p (\frac{y={s}}{x;\theta } )\right )\\
-\frac{1}{\left | \mathcal{O} \right | }\sum_{(x,y)\in \mathcal{C}^{s} }log \left ( p (\frac{y={o}}{x;\theta } )\right )
\end{split}
\end{equation}
Due to the mismatching of feature dimension, decomposed language features and image features need to be converted to consistent, the formulation is as follows:
\begin{equation}
    f^{+}_{t\to i} = \mathcal{T}_{txt2img}(f^{+}_{t}), f_{i \to t} = \mathcal{T}_{img2txt}(f_{v}),
\end{equation}
where $\mathcal{T}_{txt2img}$ is the transformation from language feature to image feature and $\mathcal{T}_{img2txt}$ is opposite.

The key to the image-text matching task is how to accurately calculate the visual-semantic similarity between images and texts. However, most of the existing algorithms only focus on the association between elements within a single modality, and do not combine the image and text features.
Since the image and text encoders are all based on transformer layers, we utilize cross-attention and self-attention mechanism to fuse the decomposed feature with image feature~\cite{wei2020multi, vaswani2017attention, chen2021crossvit}.
Let $S_{1}$ and $S_{2}$ be the two modalities to be fused, and the fusion from $S_{1}$ to $S_{2}$ can be described as follows:
\begin{equation}
    \mathcal{F}(S_{1} \to S_{2}) = softmax((W_{Q}S_{2})(W_{K}S_{1})^{T})W_{V}S_{1},
\end{equation}
where $W_{Q}$, $W_{K}$, $W_{V}$ are trainable parameters and denote the query, key and value similar to Multi-Head Self-Attention~\cite{vaswani2017attention}.
So the specific fusion of cross-attention is as follows:
\begin{equation}
\begin{split}
    f^{fused}_{v} = \mathcal{F}(f^{+}_{t\to i} \to f_{v}), \\
    f^{fused}_{t} = \mathcal{F}(f_{i \to t} \to f^{+}_t).
\end{split}
\end{equation}
To establish a deeper relationship among the fused features, they can to be fine-tuned by self-attention and the formulation is as follows.
\begin{equation}
\begin{split}
    f^{fused}_{v} = \mathcal{F}(f^{fused}_{v} \to f^{fused}_{v}), \\ 
    f^{fused}_{t} = \mathcal{F}(f^{fused}_{t} \to f^{fused}_{t}).
\end{split}
\end{equation}
Besides, the fusion module consists of cross-attention and self-attention can be repeated $K$ times to better adapted to complex tasks.
Since the language features are decomposed, they can be recomposed before comparing with image features, which is a reverse phase compared to the decomposition. The formula is as follows:
\begin{equation}
    f_{s}, f_{o} = S(f^{fused}_{t}),
    f_{t} = \left \{ MLP(f_{s}\cdot f_{o}) | (s, o) \in \mathcal{C}^{s} \right \},
\end{equation}
where $S(\cdot)$ is the split function and $MLP(\cdot)$ is the multi-layer perception to fine-tune.
Finally, the DFM class probability $p_{dfm}(\frac{y={(s,o)}}{x:\theta } )$ as follows:
\begin{equation}
    p_{dfm}(\frac{y={(s,o)}}{x;\theta } )=\frac{exp(f_{v}\cdot f_{t})}{\sum _{(\bar{s},\bar{o} ) \in \mathcal{C}^{s} }exp(f_{v} \cdot  f_{t}) } .
\end{equation}
And the corss entropy loss in DFM can be minimized:
\begin{equation}
    \mathcal{L}_{dfm}= -\frac{1}{\left | \mathcal{C}^{s} \right | }\sum_{(x,y)\in \mathcal{C}^{s} }log \left ( p_{dfm} (\frac{y={(s,o)}}{x;\theta } )\right )    .
\end{equation}
The overall loss of the framework DFSP can be summarized as follows:
\begin{equation}
    \mathcal{L} =\mathcal{L}_{dfm} + \alpha \mathcal{L}_{st+obj} + \beta \mathcal{L}_{spm},
\end{equation}
in which $\alpha$ and $\beta$ are the weighting coefficients to balance the influence of each loss.

DFSP can be divided into three categories according to the fusion methods, \textit{BiF}, \textit{i2t} and \textit{t2i}.
While \textit{i2t} and \textit{t2i} respectively mean the fusion of image feature into text feature and text feature into image feature, while \textit{BiF} means the fusion in both directions.
Fusion of image feature with text feature requires decomposition of the text feature, and to maintain joint representation in the pair space, the fused features need to be recomposed, which is reverse of decomposition.

\subsection{Inference}
We utilize the final fused probability to infer on test set, and the test set includes seen and unseen compositions, which can be denoted as $\mathcal{C}^s \cup \mathcal{C}^u$. For both closed-world and open-world settings in testing phase, the most likely predicted result can by calculated as follows:
\begin{equation}
    \hat{y} = argmax (p_{dfm}(\frac{y={(s,o)}}{x:\theta } )), y \in \mathcal{C}^s \cup \mathcal{C}^u.
\end{equation}

To filter out infeasible compositions in the open-world setting, we follow the post-training calibration method~\cite{mancini2022learning, nayak2022learning}.
First, we calculate the similarities between objects:
\begin{equation}
    q_{o}(s, o) = max\frac{\phi (o) \cdot \phi (\hat{o} )}{\left \| \phi(o) \right \| \left \| \phi (\hat{o} ) \right \| } ,(o,\hat{o})\in \mathcal{O} ,
\end{equation}
where $q_{o}(s, o)$ denotes the similarity between object $o$ and $\hat{o}$, and $\phi (\cdot)$ is an embedding function. Also, the similarities of states can be obtained in the same way.
Next, the feasibility score can be calculated by mean pooling $\mu $:
\begin{equation}
    q(s,o) = \mu (q_s(s,o), q_o(s,o)).
\end{equation}
Finally, the infeasible compositions can be filter out by a threshold T:
\begin{equation}
    \hat{y} = argmax (p(\frac{y={(s,o)}}{x:\theta } )), y \in \mathcal{C}^s \cup \mathcal{C}^u, q_(s,o)>T.
\end{equation}

\section{Experiment}
In this section, we describe all datasets and our experiments. And the comparisons with other state-of-the-art methods are presented in detail. Finally, the ablation experiments prove the efficiency of our algorithm.

\subsection{Experiment Setup}
\textbf{Datasets.} We experiment with three real-world challenging benchmark datasets: MIT -States~\cite{isola2015discovering}, UT-Zappos~\cite{yu2014fine} and C-GQA~\cite{naeem2021learning} respectively.
Specifically, MIT-States contains 53753 natural images, with 115 states and 245 objects. In the closed-world settings, the search space contains 1262 seen compositions and 300 unseen for validation and 400 unseen for test. 
UT-Zappos consists of 50025 images of shoes, with 16 states and 12 objects. For the closed-world experiments, it is constrained to the 83 seen and 15/18 (validation/test) unseen compositions. And for UT-Zappos, we follow the split in~\cite{purushwalkam2019task}. For about C-GQA, the most pairs dataset for CZSL, contains 453 states and 870 objects, with 39298 images in total, which contains over 9500 compositions. Finally, in the open-world settings, these datasets contain 28175, 192 and 278362 respectively.

\textbf{Metrics.}
Following the setting of prior work~\cite{mancini2021open}, we compute the prediction accuracy based on the seen and unseen compositions both in the closed-world and open-world scenarios. Specifically, \textit{Seen (S)} denotes the accuracy tested only on seen compositions and \textit{Unseen (U)} represents the accuracy evaluated only on unseen compositions. Also, we can calculate \textit{Harmonic Mean (HM)} of the \textit{S} and \textit{U} metrics. 
Since zero-shot models have inherent bias for seen compositions, we can draw a seen-unseen accuracy curve at different operating points with the bias from $-\infty $ to $+\infty$ to compute the \textit{Area Under the Curve (AUC)}.
To sum up, the metrics consist of \textit{S}, \textit{U}, \textit{HM} and \textit{AUC}.

\textbf{Implementation Details.}
We implement DFSP with PyToch 1.12.1~\cite{paszke2019pytorch} and optimized by Adam optimizer over the three challenging datasets for 20 epochs. The image encoder and text encoder are both based on the pretrained CLIP Vit-L/14 model, and the entire model are trained and evaluated on 1$\times$NVIDIA RTX 3090 GPU.
Besides, we set the number of fusion blocks $K$ and self-attention section as 1, and the evaluation metrics are tested on the model which computes lowest loss during the validation phase.

\begin{table*}[!htb]\centering
\caption{Closed-world results on MIT-States, UT-Zappos and C-GQA. \textit{S} and \textit{U} are the predict accuracies evaluated on seen and unseen compositions. \textit{H} is the harmonic mean of \textit{U} and \textit{S} and \textit{AUC} is the area under the curve. The best results are in bold.} 
\label{tab:closed-world}
\resizebox{.88\textwidth}{!}{
\begin{tabular}{ccccclcccclcccc}
\hline
\multirow{2}{*}{Method} & \multicolumn{4}{c}{MIT-States} &  & \multicolumn{4}{c}{UT-Zappos} &  & \multicolumn{4}{c}{CGQA} \\ \cline{2-5} \cline{7-10} \cline{12-15} 
                        & S      & U     & H     & AUC   &  & S     & U     & H     & AUC   &  & S    & U    & H    & AUC \\ \hline
AoP~\cite{nagarajan2018attributes}                     & 14.3   & 17.4  & 9.9   & 1.6   &  & 59.8  & 54.2  & 40.8  & 25.9  &  & 17.0 & 5.6  & 5.9  & 0.7 \\
LE+~\cite{naeem2021learning}                     & 15.0   & 20.1  & 10.7  & 2.0   &  & 53.0  & 61.9  & 41.0  & 25.7  &  & 18.1 & 5.6  & 6.1  & 0.8 \\
TMN~\cite{purushwalkam2019task}                     & 20.2   & 20.1  & 13.0  & 2.9   &  & 58.7  & 60.0  & 45.0  & 29.3  &  & 23.1 & 6.5  & 7.5  & 1.1 \\
SymNet~\cite{li2020symmetry}                  & 24.2   & 25.2  & 16.1  & 3.0   &  & 49.8  & 57.4  & 40.4  & 23.4  &  & 26.8 & 10.3 & 11.0 & 2.1 \\
CompCos~\cite{mancini2021open}                 & 25.3   & 24.6  & 16.4  & 4.5   &  & 59.8  & 62.5  & 43.1  & 28.1  &  & 28.1 & 11.2 & 12.4 & 2.6 \\
CGE~\cite{naeem2021learning}                     & 28.7   & 25.3  & 17.2  & 5.1   &  & 56.8  & 63.6  & 41.2  & 26.4  &  & 28.7 & 25.3 & 17.2 & 5.1 \\
Co-CGE~\cite{mancini2022learning}                  & 31.1   & 5.8   & 6.4   & 1.1   &  & 62.0  & 44.3  & 40.3  & 23.1  &  & 32.1 & 2.0  & 3.4  & 0.5 \\
SCEN~\cite{li2022siamese}                    & 29.9   & 25.2  & 18.4  & 5.3   &  & 63.5  & 63.1  & 47.8  & 32.0  &  & 28.9 & 25.4 & 17.5 & 5.5 \\
CSP~\cite{nayak2022learning}                     & 46.6   & 49.9  & 36.3  & 19.4  &  & 64.2  & 66.2  & 46.6  & 33.0  &  & 28.8 & 26.8 & 20.5 & 6.2 \\ \hline
\textbf{DFSP}(\textit{i2t})     & \textbf{47.4}   & 52.4  & 37.2  & 20.7  &  & 64.2  & 66.4  & 45.1  & 32.1  &  & 35.6 & 29.3 & 24.3 & 8.7 \\ 
\textbf{DFSP}(\textit{BiF})     & 47.1   & \textbf{52.8}   &\textbf{ 37.7}  & \textbf{20.8}  &  & 63.3  & 69.2  & 47.1  & 33.5  &  & 36.5 & \textbf{32.0} & 26.2 & 9.9 \\ 
\textbf{DFSP}(\textit{t2i})     & 46.9   & 52.0  & 37.3  & 20.6  &  & \textbf{66.7}  & \textbf{71.7}  & \textbf{47.2}  & \textbf{36.0}  &  & \textbf{38.2} & \textbf{32.0} & \textbf{27.1} & \textbf{10.5} \\ \hline
\end{tabular}}
\end{table*}

\begin{figure*}[t]
  \centering
  \includegraphics[scale=0.48]{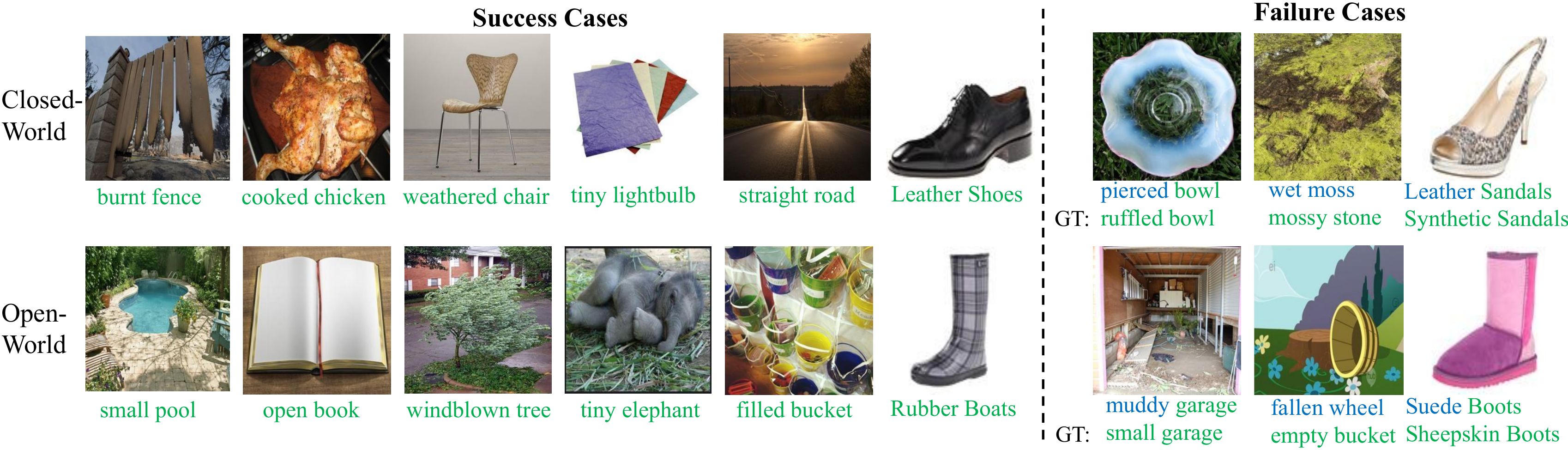}
   \caption{Qualitative results. We evaluate top-1 predictions for some cases on MIT-States and UT-Zappos. The first row shows the results of the closed-world and the bottom row is the open-world. Six cols on the left are examples of successful predictions, and three on the right are examples of failures. For the failure cases, \textcolor{blue}{blue} denotes the wrong prediction and all images are randomly selected.}
   \label{fig:demo}
\end{figure*}

\subsection{Comparision with State-of-the-Arts}
Experimental comparisons with the prior compositional zero-shot learning methods are reported, including  AoP~\cite{nagarajan2018attributes}, LE+~\cite{naeem2021learning}, TMN~\cite{purushwalkam2019task}, SymNet~\cite{li2020symmetry}, CompCos~\cite{mancini2021open}, CGE~\cite{naeem2021learning}, Co-CGE~\cite{mancini2022learning}, SCEN~\cite{li2022siamese}, KG-SP~\cite{karthik2022kg} and recent proposed CSP~\cite{nayak2022learning}. For our proposed method DFSP, we test a variety of different fusion methods: \textit{BiF}, \textit{i2t} and \textit{t2i}, which denotes the fusion direction as bidirectional fusion, fusion on the text branch and fusion on the image branch. The experiment is conducted on both closed-world and open-world, and the results are shown in Tab. \ref{tab:closed-world} and Tab. \ref{tab:open-world}.

For the closed-world setting, DFSP in Tab. \ref{tab:closed-world} shows that our method achieves the new state-of-the-art on MIT-States, UT-Zappos and CGQA datasets. DFSP reaches the highest \textit{AUC} of 20.8\% on MIT-States, 36.0\% on UT-Zappos and 10.5\% on CGQA, which outperforms CSP by 4.3\%.  
Besides, we improve the harmonic mean by 6.6\% on CGQA relative to other existing methods.
And the seen and unseen accuracies on these datasets are also the best results.

Tab. \ref{tab:open-world} shows the DFSP results on open-world setting and we also get the best results on all metrics. During the inference stage, the infeasible filter threshold $T$ is fixed on 0.4, and DFSP outperforms CSP by 1.1\% on MIT-States and KG-SP by 3.8\% on UT-Zappos for the \textit{AUC} metric.
It can be clearly seen that the unseen accuracy in open-world has improved a lot like 15.9\% on UT-Zappos, which proves that the decomposes state and object features fused with the image feature can really enhance the sensitiveness for unseen compositions. This improvement can also drive the \textit{H} metric to the best.

Combining the results of closed-world and open-world, it can be seen that the fusion method of \textit{t2i} is the best, showing that fused with decomposed features on the image branch can achieve better performance than that on the language branch due to the destruction of joint representation even they can be recomposed.
Experimental results on three challenging datasets demonstrate that our proposed Decomposed Fusion with Soft Prompt framework (DFSP) can effectively improve the performance of the model for compositional zero-shot learning.

\begin{table*}[!htb]\centering
\caption{Open-world results on MIT-States, UT-Zappos and C-GQA. \textit{S} and \textit{U} are the predict accuracies evaluated on seen and unseen compositions. \textit{H} is the harmonic mean of \textit{U} and \textit{S} and \textit{AUC} is the area under the curve. The best results are in bold.} 
\label{tab:open-world}
\resizebox{.88\textwidth}{!}{
\begin{tabular}{ccccclcccclcccc}
\hline
\multirow{2}{*}{Method}        & \multicolumn{4}{c}{MIT-States} &  & \multicolumn{4}{c}{UT-Zappos} &  & \multicolumn{4}{c}{CGQA} \\ \cline{2-5} \cline{7-10} \cline{12-15} 
                               & S      & U      & H     & AUC  &  & S     & U     & H     & AUC   &  & S     & U   & H   & AUC  \\ \hline
AoP~\cite{nagarajan2018attributes}                            & 16.6   & 5.7    & 4.7   & 0.7  &  & 50.9  & 34.2  & 29.4  & 13.7  &  & -     & -   & -   & -    \\
LE+~\cite{naeem2021learning}                            & 14.2   & 2.5    & 2.7   & 0.3  &  & 60.4  & 36.5  & 30.5  & 16.3  &  & 19.2  & 0.7 & 1.0 & 0.08 \\
TMN~\cite{purushwalkam2019task}                            & 12.6   & 0.9    & 1.2   & 0.1  &  & 55.9  & 18.1  & 21.7  & 8.4   &  & -     & -   & -   & -    \\
SymNet~\cite{li2020symmetry}                         & 21.4   & 7.0    & 5.8   & 0.8  &  & 53.3  & 44.6  & 34.5  & 18.5  &  & 26.7  & 2.2 & 3.3 & 0.43 \\
CompCos~\cite{mancini2021open}                        & 25.4   & 10.0   & 8.9   & 1.6  &  & 59.3  & 46.8  & 36.9  & 21.3  &  & -     & -   & -   & -    \\
CGE~\cite{naeem2021learning}                            & 32.4   & 5.1    & 6.0   & 1.0  &  & 61.7  & 47.7  & 39.0  & 23.1  &  & 32.7  & 1.8 & 2.9 & 0.47 \\
Co-CGE\textasciicircum{}Closed~\cite{mancini2022learning} & 31.1   & 5.8    & 6.4   & 1.1  &  & 62.0  & 44.3  & 40.3  & 23.1  &  & 32.1  & 2.0 & 3.4 & 0.53 \\
Co-CGE\textasciicircum{}Open~\cite{mancini2022learning}   & 30.3   & 11.2   & 10.7  & 2.3  &  & 61.2  & 45.8  & 40.8  & 23.3  &  & 32.1  & 3.0 & 4.8 & 0.78 \\
KG-SP~\cite{karthik2022kg}                          & 28.4   & 7.5    & 7.4   & 1.3  &  & 61.8  & 52.1  & 42.3  & 26.5  &  & 31.5  & 2.9 & 4.7 & 0.78 \\
CSP~\cite{nayak2022learning}                            & 46.3   & 15.7   & 17.4  & 5.7  &  & 64.1  & 44.1  & 38.9  & 22.7  &  & 28.7  & 5.2 & 6.9 & 1.20 \\ \hline
\textbf{DFSP}(\textit{i2t})            & 47.2   & 18.2   & 19.1  & 6.7  &  & 64.3  & 53.8  & 41.2  & 26.4   &  & 35.6  & 6.5 & 9.0 & 1.95 \\ 
\textbf{DFSP}(\textit{BiF})            & 47.1   & 18.1   & 19.2  & 6.7  &  & 63.5  & 57.2  & 42.7  & 27.6  &  & 36.4   & \textbf{7.6} & \textbf{10.6} & 2.39 \\ 
\textbf{DFSP}(\textit{t2i})            & \textbf{47.5}   & \textbf{18.5}   & \textbf{19.3}  & \textbf{6.8}  &  & \textbf{66.8}  & \textbf{60.0}  & \textbf{44.0}  & \textbf{30.3}  &  & \textbf{38.3}  & 7.2 & 10.4 & \textbf{2.40} \\ \hline
\end{tabular}}
\end{table*}

\begin{table}[htb]\centering
\caption{Ablation study experiments on MIT-States and UT-Zappos with the setting of closed-world (CW) and open-world (OW). The best results are in bold.} 
\label{tab:ablation}
\resizebox{.48\textwidth}{!}{
\begin{tabular}{ccccccccccc}
\hline
\multicolumn{2}{c}{\multirow{2}{*}{Method}}                                 & \multicolumn{4}{c}{MIT-States}                                &                      & \multicolumn{4}{c}{UT-Zappos}                                 \\ \cline{3-6} \cline{8-11} 
\multicolumn{2}{c}{}                                                        & S             & U             & H             & AUC           &                      & S             & U             & H             & AUC           \\ \hline
\multicolumn{1}{c|}{\multirow{4}{*}{CW}} & \textit{SPM}                              & 45.8          & 50.2          & 35.8          & 19.1          &                      & 64.6          & 63.9          & 47.0          & 33.1          \\
\multicolumn{1}{c|}{}                    & \multicolumn{1}{l}{\textit{+SA}} & 47.6          & 51.9          & \textbf{37.4} & 20.6          &                      & 65.3         & 66.2           & 46.6          & 32.6           \\
\multicolumn{1}{c|}{}                    & \textit{+Fusion}                    & 46.9 & 47.5          & 34.9          & 18.3 &                      & 62.9          & 41.4          & 38.0          & 21.4          \\
\multicolumn{1}{c|}{}                    & \textit{+DeC}                    & \textbf{48.3} & 51.3          & 37.3          & \textbf{20.7} &                      & 62.3          & 70.7          & 46.8          & 33.5          \\
\multicolumn{1}{c|}{}                    & \textit{+DFM}                    & 46.9          & \textbf{52.0} & 37.3          & 20.6          &                      & \textbf{66.7} & \textbf{71.7} & \textbf{47.2} & \textbf{36.0} \\ \hline
\multicolumn{1}{c|}{\multirow{4}{*}{OW}} & \textit{SPM}                     & 45.8          & 16.7          & 18.3          & 6.1           &                      & 64.6          & 44.6          & 40.8          & 23.5          \\
\multicolumn{1}{c|}{}                    & \multicolumn{1}{l}{\textit{+SA}} & \textbf{47.6} & 17.6          & 19.0          & 6.6           &                      & 65.4          & 54.6          & 43.0          & 26.9          \\
\multicolumn{1}{c|}{}                    & \textit{+DeC}                    & 47.4          & 17.7          & 19.3          & 6.6           &                      & 61.7          & 57.1          & 42.3          & 26.4          \\
\multicolumn{1}{c|}{}                    & \textit{+DFM}                    & 47.5          & \textbf{18.5} & \textbf{19.3} & \textbf{6.8}  & \multicolumn{1}{l}{} & \textbf{66.8} & \textbf{60.0} & \textbf{44.0} & \textbf{30.3} \\ \hline
\end{tabular}}
\end{table}

\subsection{Ablation Study}
To evaluate the effectiveness of DFSP, we also establish an ablation study on MIT-States and UT-Zappos. The soft prompt module in Fig. \ref{fig:framework} is the base recognition model and the DFSP version is \textit{t2i}.
Meanwhile, we evaluate models with only self-attention (\textit{SA}), with only fusion (\textit{Fusion}), with only decomposition section (\textit{DeC}) and with decomposed fusion module (\textit{DFM}).
The closed-world and open-world experimental results can be seen in Tab. \ref{tab:ablation}.

\textbf{Effectiveness of SPM.}
The experimental results show that the model with only \textit{SPM} can improve a little compared with CSP, demonstrating its fully learnable soft prompts can be better adapted to downstream supervised tasks. 
If only the language feature is decomposed and not integrated, the experimental results can be improved a lot in multiple metrics, including closed-world and open-world.
Meanwhile, the results of \textit{+SA} are also significantly improved compared to \textit{SPM}, which proves the fine-tuning effect of adding self-attention to the model, making it better to transfer VLMs to new tasks.

\textbf{Effectiveness of DFM.}
Eventually, the results of +\textit{DFM} show a very large improvement, proving the effectiveness of DFSP.
However, the results of \textit{+SA} and \textit{+DeC} are not much different, and \textit{+DFM} both decomposes language features and fuses with image features, which will lead to a particularly large improvement.
Also, only \textit{+Fusion} will make the effect worse, mainly causing increasing the bias of seen compositions, which does not meet the definition of CZSL.
To a certain extent, this shows that decomposition and fusion complement each other, and only decomposition in the form of combination is the same as the essence of \textit{+SA}.
From the results of metric \textit{U}, DFSP can also demonstrate the high response of DFM to unseen compositions.
Besides, this is not limited to closed-world, open-world has also seen significant improvements.

\subsection{Qualitative Results}
We report qualitative results for seen and unseen compositions with top-1 predictions both on closed-world and open-world in Fig. \ref{fig:demo}.
Our model can really be generalized well to unseen compositions, alleviating the domain gap between seen and unseen sets. 
Meanwhile, benefited from the joint prompt consists of state and object, DFSP can predict the compositions with high accuracy.
For the failure cases, the most prone to error is the prediction of state, but even if the prediction is not correct, it still conforms to the combination logic of state and object.
Great results can be seen on both closed-world and open-world setting, indicating that the model is not restricted by open-world scenario.


\subsection{Why DFSP can work well?}
Extensive experiments show the efficiency of DFSP and we analysis the reasons of this.
Firstly, since the encoders of DFSP have been pretrained on a large-scale image-text pairs dataset, the model can be really fine-tuned well with a targeted prompt like CSP.
Compared with CSP, there are more parameters can be fine-tuned in DFSP to be adapted to new supervised tasks~\cite{gao2021clip}.
With the Decomposed Fusion Module (DFM), the pair space is transformed from the original pairing of "language" and "image" to the pairing of "language + (image)" and "(state and object) + image" (as shown in Fig. \ref{fig:framework}), and both decomposed state feature and object feature can respond to image feature in the fusion stage. 
Being fused with the decomposed features, image feature can establish its relation to state and object, then pair with another branch (like language feature in DFSP (\textit{t2i})) to guide the results beyond the seen compositions during training. 
Benefited from this, the overall model can be more sensitive to the unseen compositions in the pair space.

\section{Conclusion}
In this work, we propose a novel framework termed Decomposed Fusion with Soft Prompt (DFSP) to effectively recognize the unknown compositions of state and object during training.
Based on the vision-language paradigm, we firstly establish a learnable soft prompt consists of prefix, state and object to construct the joint representation of state and object.
Besides, we design a Decomposed Fusion Module (DFM) to fuse the language features with image features, which can enable cross-modal interactions between them.
Meanwhile, DFM decomposes the language feature to unattached state and object features, and then they will be fused with image feature to guide enhancing fusion.
Benefited from DFM, image feature could learn relations with state and object features, which improves the response of unseen compositions in the pair space.
Extensive experiments on three challenging datasets demonstrate the efficiency of our proposed method DFSP.

\newpage

{\small
\bibliographystyle{ieee_fullname}
\bibliography{egbib}
}

\newpage

\end{document}


\maketitle

The supplementary file provides more details for paper "Decomposed Soft Prompt Guided Fusion Enhancing for Compositional Zero-Shot Learning", including the following aspects:
\begin{enumerate}
    \item Summary Statistics of Datasets.
    \item Backbone Study.
    \item Hyper-Parameters Analysis.
    \item Pseudocode.
    \item Qualitative Results.
\end{enumerate}

\section{Summary Statistics of Datasets}
We analyse three datasets included MIT -States~\cite{isola2015discovering}, UT-Zappos~\cite{yu2014fine} and C-GQA~\cite{naeem2021learning} statistically and the summary statistics can be seen in Tab.~\ref{tab:sum_data}. $s$ and $o$ denote the number of \textit{state} and \textit{object} concepts, and $i$ represents the number of images. Also, $c_s$ and $c_u$ are the pair concepts of seen and unseen classes.

\begin{table}[htb]
\caption{Summary statistics of the datasets used in our experiments, including MIT-States, UT-Zappos and CGQA.}
\centering
\resizebox{.8\textwidth}{!}{
\begin{tabular}{ccc|cc|ccc|ccc}
\hline
\multicolumn{3}{c|}{}                         & \multicolumn{2}{c|}{\textbf{Train}} & \multicolumn{3}{c|}{\textbf{Validation}} & \multicolumn{3}{c}{\textbf{Test}}      \\
\textbf{Dataset}    & s & o & $c_s$       & i      & $c_s$  & $c_u$  & i & $c_s$ & $c_u$ & i \\ \hline
\textbf{MIT-States} & 115        & 245        & 1262              & 30338           & 300          & 300          & 10420      & 400         & 400         & 12995      \\
\textbf{UT-Zappos}  & 16         & 12         & 83                & 22998           & 15           & 15           & 3214       & 18          & 18          & 2914       \\
\textbf{CGQA}       & 453        & 870        & 6963              & 26920           & 1173         & 1368         & 7280       & 1022        & 1047        & 5098       \\ \hline
\end{tabular}}
\label{tab:sum_data}
\end{table}

\section{Backbone Study}
To evaluate the performance of DFSP with various backbone, we also retrain the model with same parameters ($\alpha=0.01$, $\beta=0.1$ and $K=1$) and only replace the backbone with ViT-B/32, Vit-B/16 and ViT-L/14@336px due to the limitation of image encoder in DFSP which is based on transformer networks~\cite{dosovitskiy2020image}. Metrics contists of \textit{S}, \textit{U}, \textit{H} and \textit{AUC} and the model (\textit{DFSP(t2i)}) are tested on MIT-States and UT-Zappos with the settings of Closed-World (CW) and Open-World (OW). 
Meanwhile, we evaluate DFSP with 5 random seeds to report the standard error, which can be seen in Tab.~\ref{tab:backbone}.

From the results in Tab.~\ref{tab:backbone}, we can see that ViT-14/L and ViT-L/14@336px achieve state-of-the-art (SOTA) results both on Closed-World and Open-World. Meanwhile, all backbones work well on multiple metrics, especially on MIT-States, such as \textit{AUC} \%12.8 versus \%5.5 on SCEN~\cite{li2022siamese} with the setting of Closed-World.

\begin{table*}[htb]
\caption{DFSP with different backbone results on MIT-States and UT-Zappos with the settings of Closed-World (CW) and Open-World (OW). We also report the standard error with 5 random seeds.}
\centering
\resizebox{.88\textwidth}{!}{
\begin{tabular}{cccccccccc}
\hline
\multicolumn{1}{l}{} &                                       & \multicolumn{4}{c}{MIT-States} & \multicolumn{4}{c}{UT-Zappos} \\
\multicolumn{1}{l}{} & Backbone                              & S     & U     & H     & AUC    & S     & U     & H     & AUC   \\ \hline
                     & ViT-B/32       & 36.7$\pm$0.25      & 43.4$\pm$0.37      & 29.4$\pm$0.18      & 13.2$\pm$0.69       & 55.8\pm1.66  & 59.5\pm2.25  & 38.5\pm2.72  & 23.3\pm1.32  \\
                     & ViT-B/16       & 39.6$\pm$0.17      & 46.5$\pm$0.28      & 31.5$\pm$0.18      & 15.1$\pm$0.32        & 62.1\pm2.20  & 67.7\pm3.21  & 48.0\pm0.89  & 33.2\pm1.21  \\
                     & ViT-L/14                              &46.8$\pm$0.54        & 52.2$\pm$0.17      & 37.4$\pm$0.35      & 20.6$\pm$0.28       & 65.2\pm1.80  & 70.7\pm1.51  & 50.4\pm2.14  & 36.8\pm0.72  \\
\multirow{-4}{*}{CW} & ViT-L/14@336px & 45.6$\pm$0.21      & 50.8$\pm$0.38      & 36.9$\pm$0.57      &  19.9$\pm$0.32      & 64.6\pm1.39  & 70.6\pm1.35  & 48.3\pm1.89  & 36.9\pm0.65  \\ \hline
                     & ViT-B/32                              &  36.7$\pm$0.19     & 12.8$\pm$0.13      & 13.1$\pm$0.25      & 3.3$\pm$0.12       & 56.8$\pm$1.19  & 38.9$\pm$2.11  & 33.8$\pm$1.30  & 17.0$\pm$0.84  \\
                     & ViT-B/16                              &  39.6$\pm$0.11     & 15.2$\pm$0.16      & 15.4$\pm$0.09      & 4.4$\pm$0.14       & 62.9$\pm$1.30  & 48.4$\pm$1.88  & 42.9$\pm$2.02  & 25.4$\pm$0.93  \\
                     & ViT-L/14                              & 47.6$\pm$0.21      &  18.7$\pm$0.35     & 19.3$\pm$0.09      &  6.7$\pm$0.13      & 64.8\pm1.22  & 59.8\pm2.19  & 45.4\pm1.21  & 29.6\pm0.88  \\
\multirow{-4}{*}{OW} & ViT-L/14@336px                        & 45.5$\pm$0.14      &  17.1$\pm$0.86     & 18.4$\pm$0.11      & 6.3$\pm$0.16       & 63.6\pm2.10  & 58.4\pm1.12  & 45.2\pm0.88  & 28.3\pm0.56  \\ \hline
\end{tabular}
}
\label{tab:backbone}
\end{table*}

\section{Hyper-Parameters Analysis}
\begin{figure}[htbp]
	\centering
	\begin{subfigure}{0.24\linewidth}
		\centering
		\includegraphics[width=0.95\linewidth]{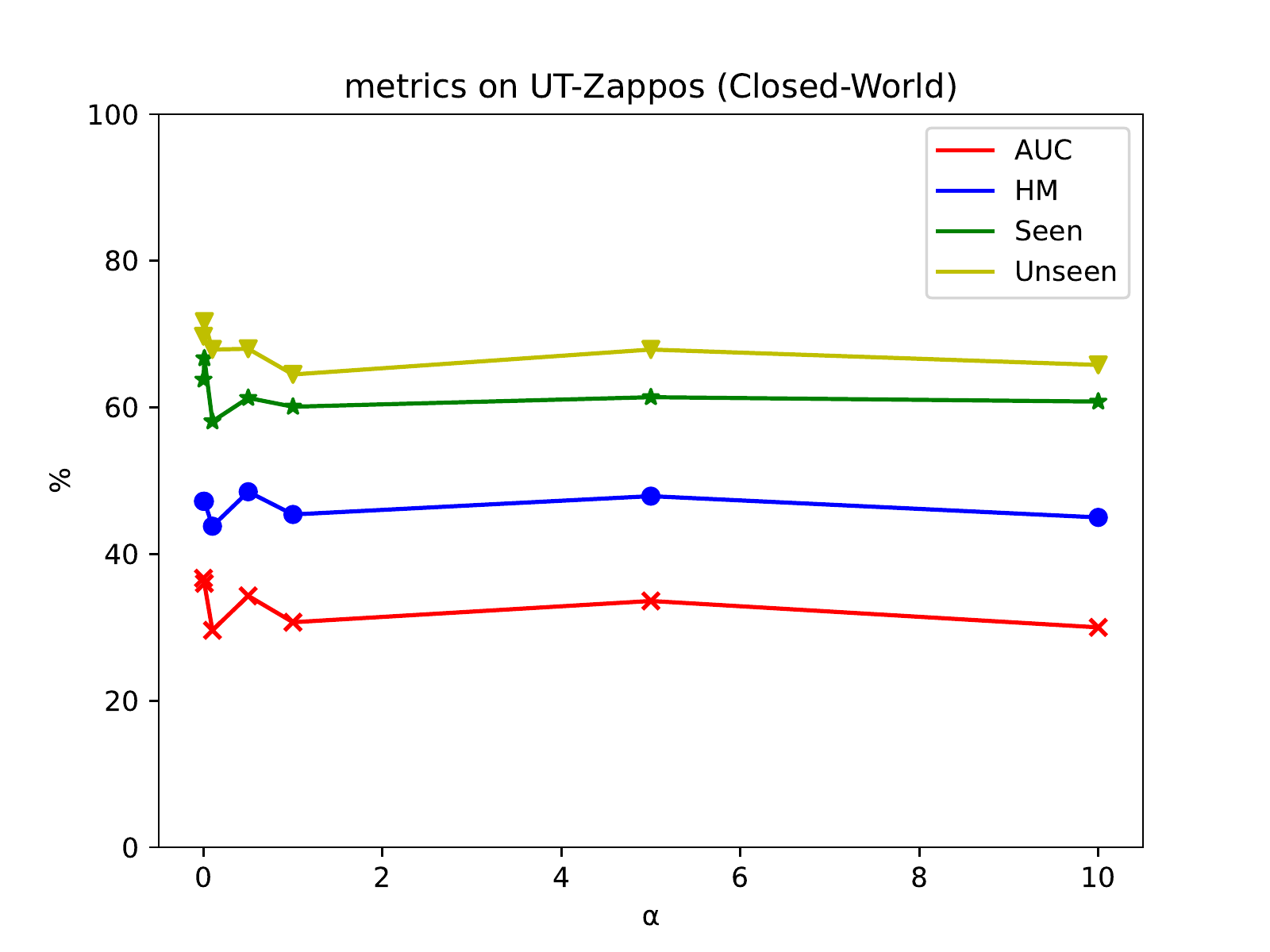}
		\caption{Closed-World ($\alpha$)}
		\label{me_ut_a_cw}
	\end{subfigure}
	\centering
	\begin{subfigure}{0.24\linewidth}
		\centering
		\includegraphics[width=0.95\linewidth]{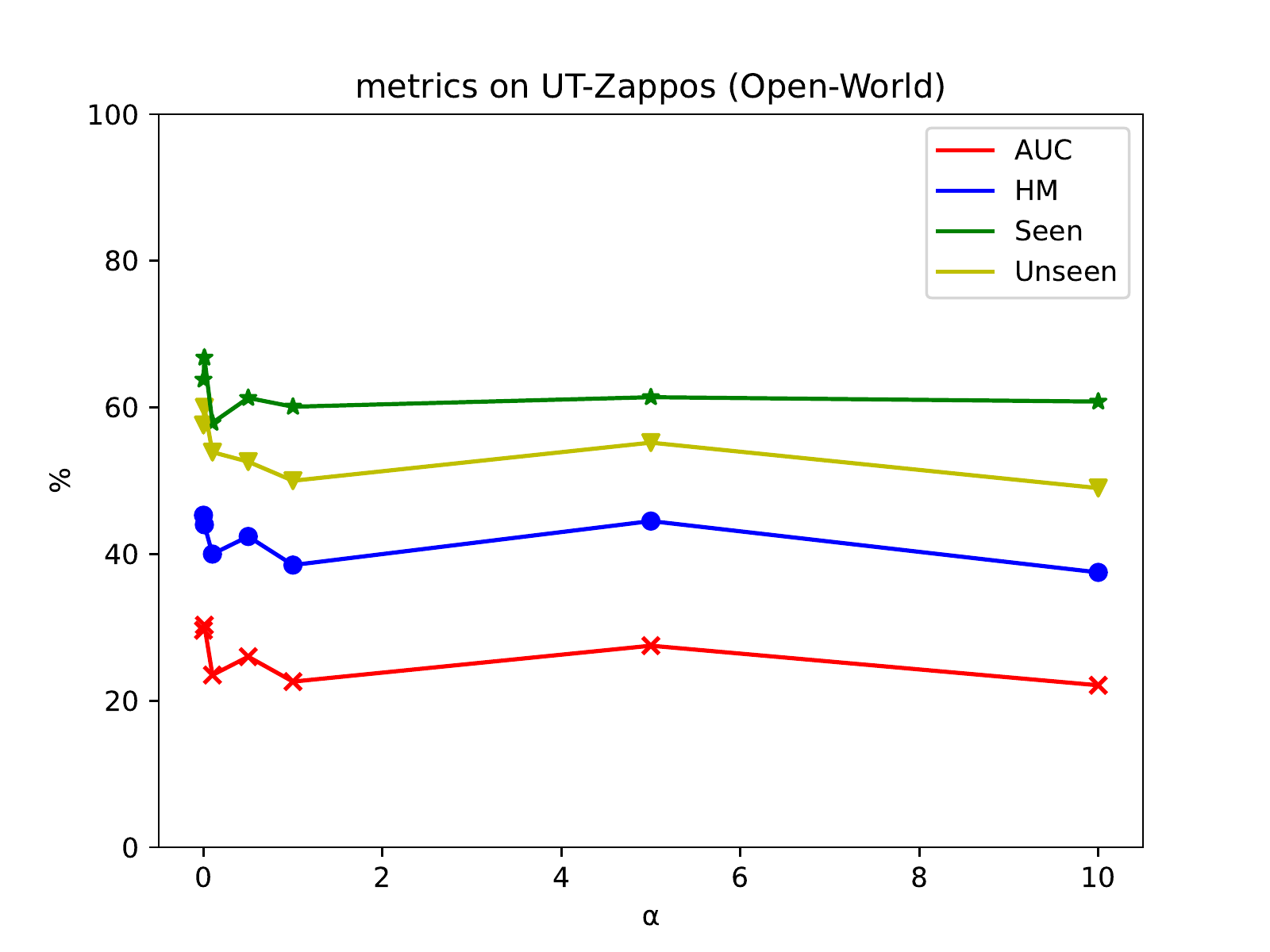}
		\caption{Open-World ($\alpha$)}
		\label{me_ut_a_ow}
	\end{subfigure}    
    \centering
	\begin{subfigure}{0.24\linewidth}
		\centering
		\includegraphics[width=0.95\linewidth]{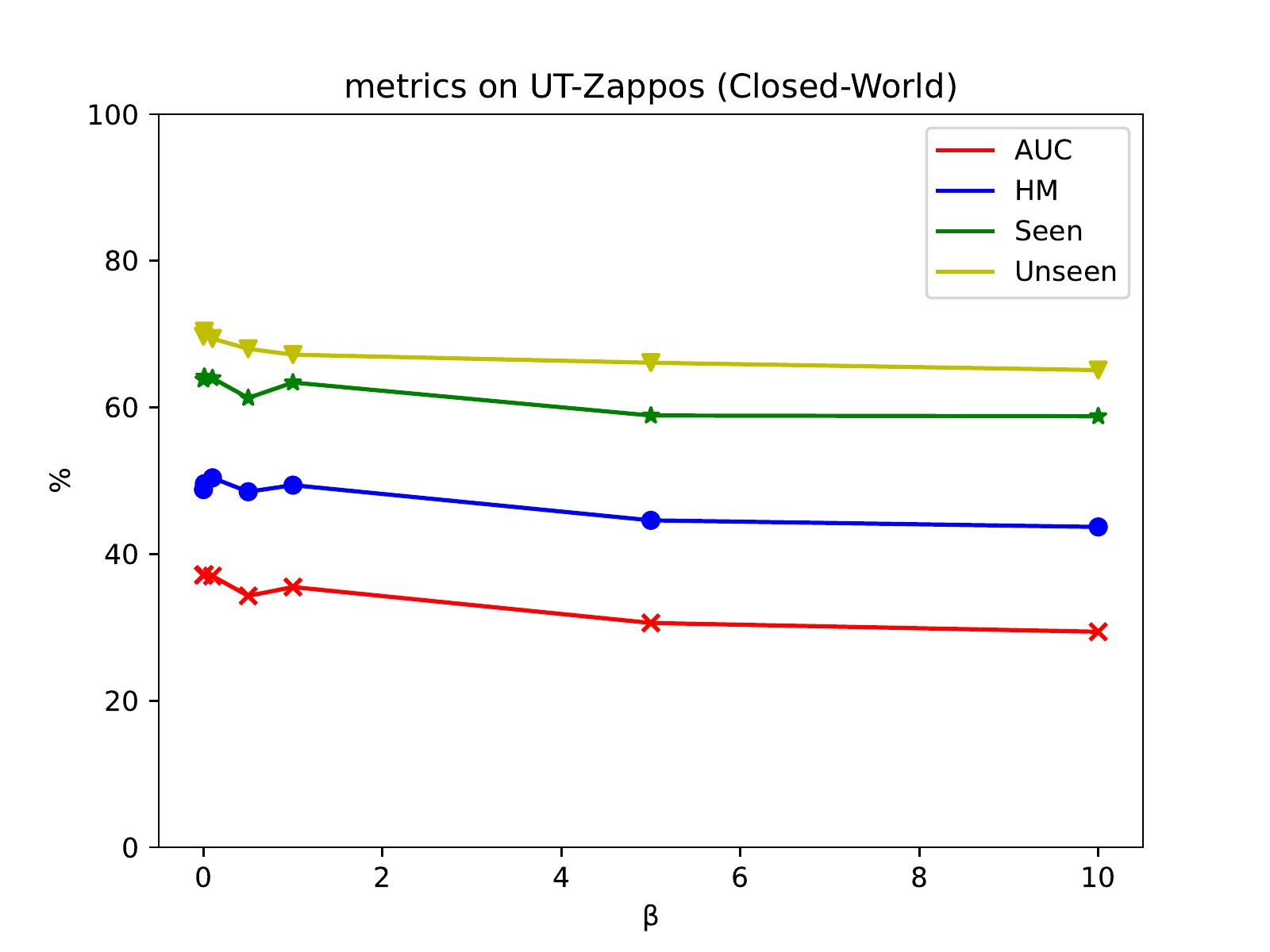}
		\caption{Closed-World ($\beta$)}
		\label{me_ut_b_cw}
	\end{subfigure}
    \begin{subfigure}{0.24\linewidth}
		\centering
		\includegraphics[width=0.95\linewidth]{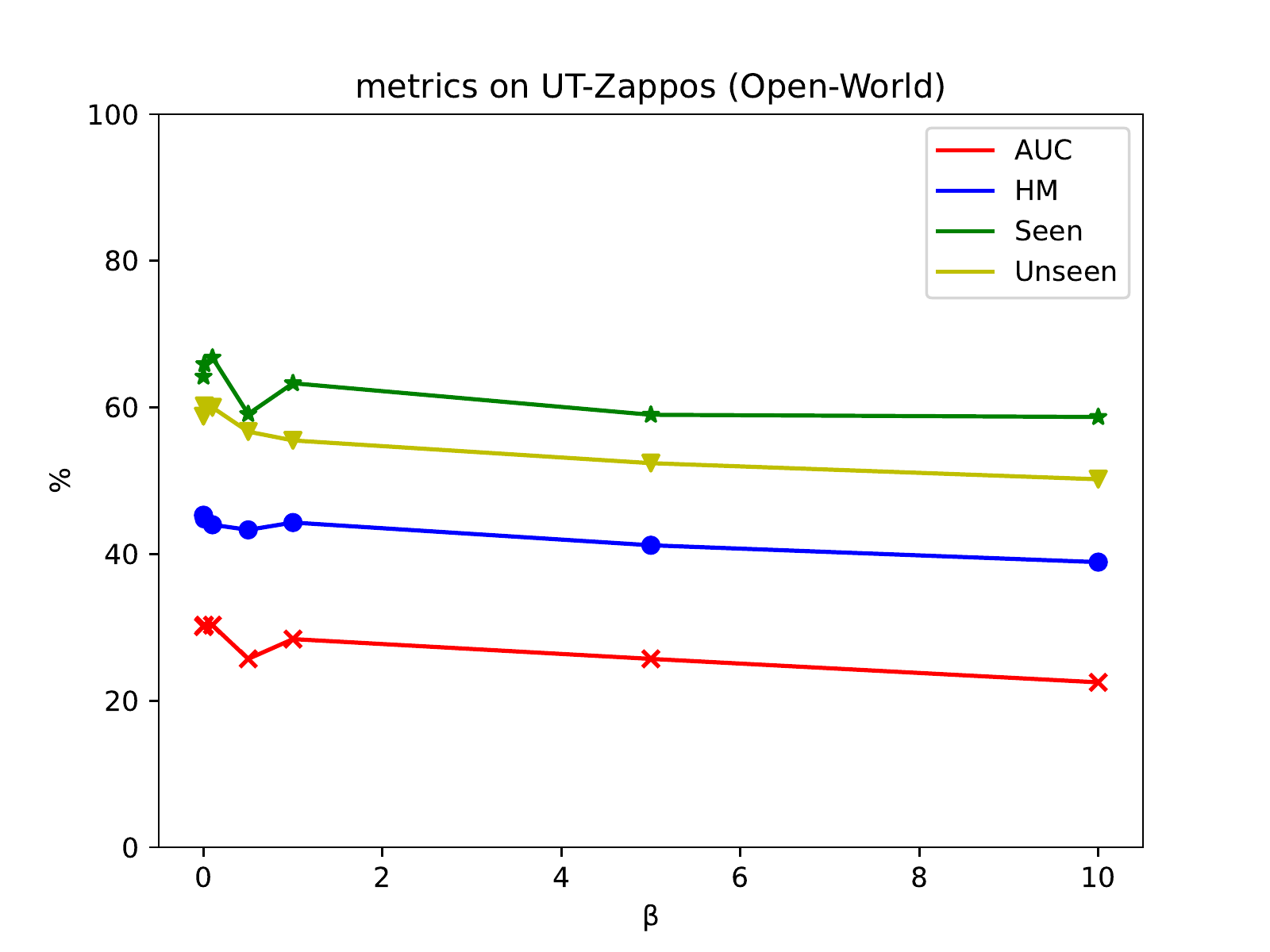}
		\caption{Open-World ($\beta$)}
		\label{me_ut_b_ow}
	\end{subfigure}
	\caption{Metrics on UT-Zappos with the settings of Closed-World and Open-World.}
	\label{fig:me_ut}
\end{figure}

\begin{figure}[htbp]
	\centering
	\begin{subfigure}{0.24\linewidth}
		\centering
		\includegraphics[width=0.95\linewidth]{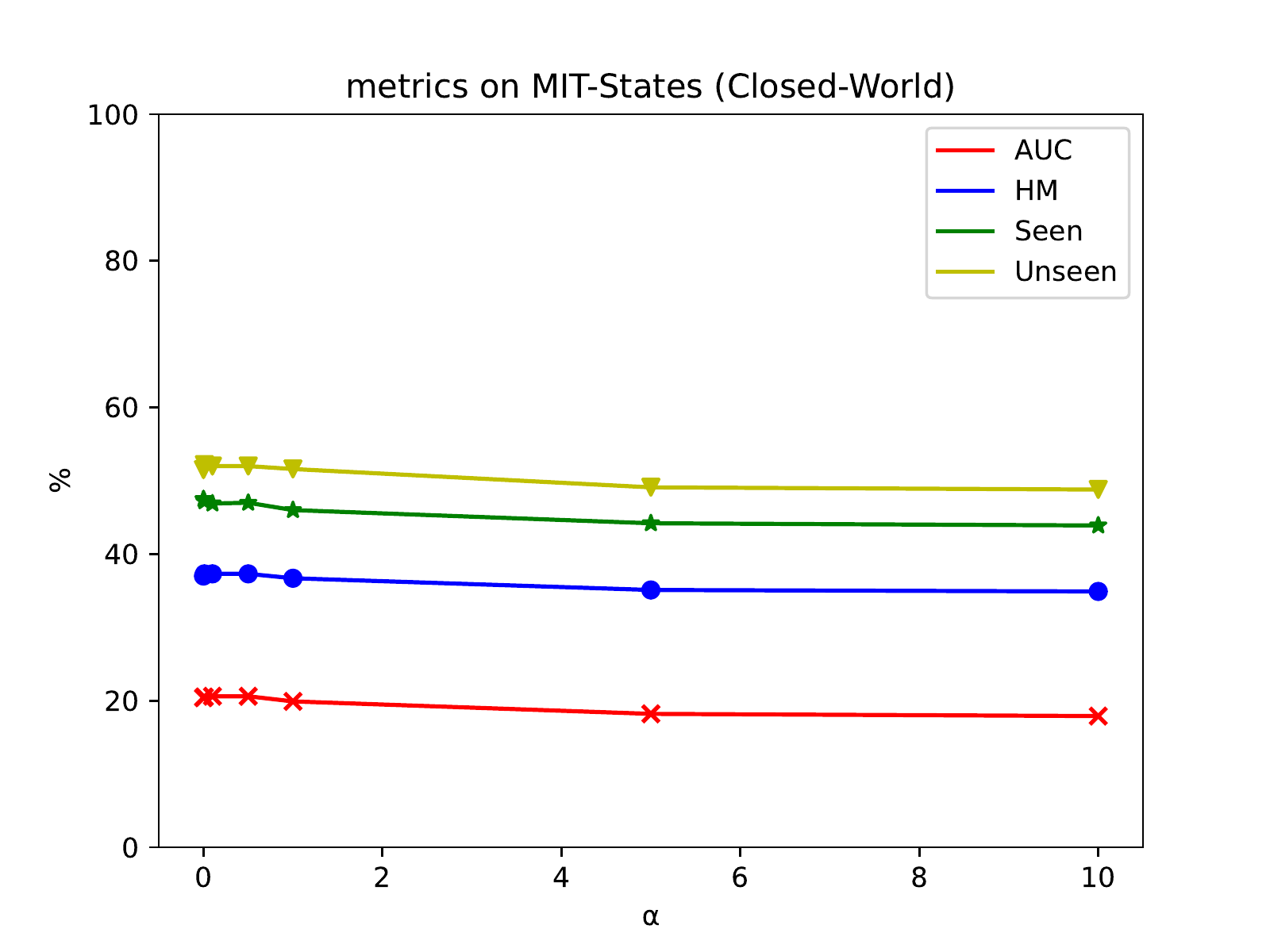}
		\caption{Closed-World ($\alpha$)}
		\label{me_mit_a_cw}
	\end{subfigure}
	\centering
	\begin{subfigure}{0.24\linewidth}
		\centering
		\includegraphics[width=0.95\linewidth]{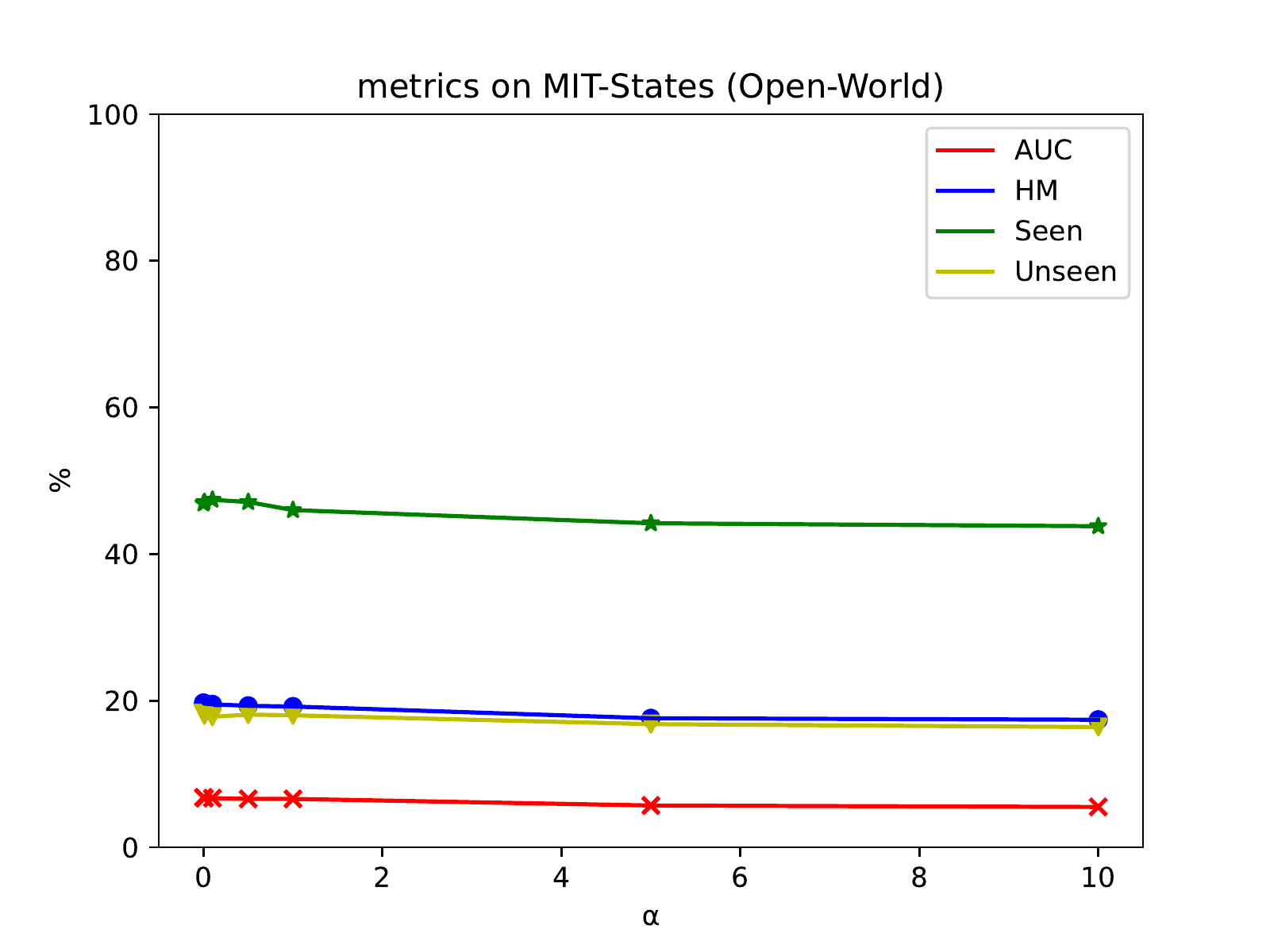}
		\caption{Open-World ($\alpha$)}
		\label{me_ut_a_ow}
	\end{subfigure}    
    \centering
	\begin{subfigure}{0.24\linewidth}
		\centering
		\includegraphics[width=0.95\linewidth]{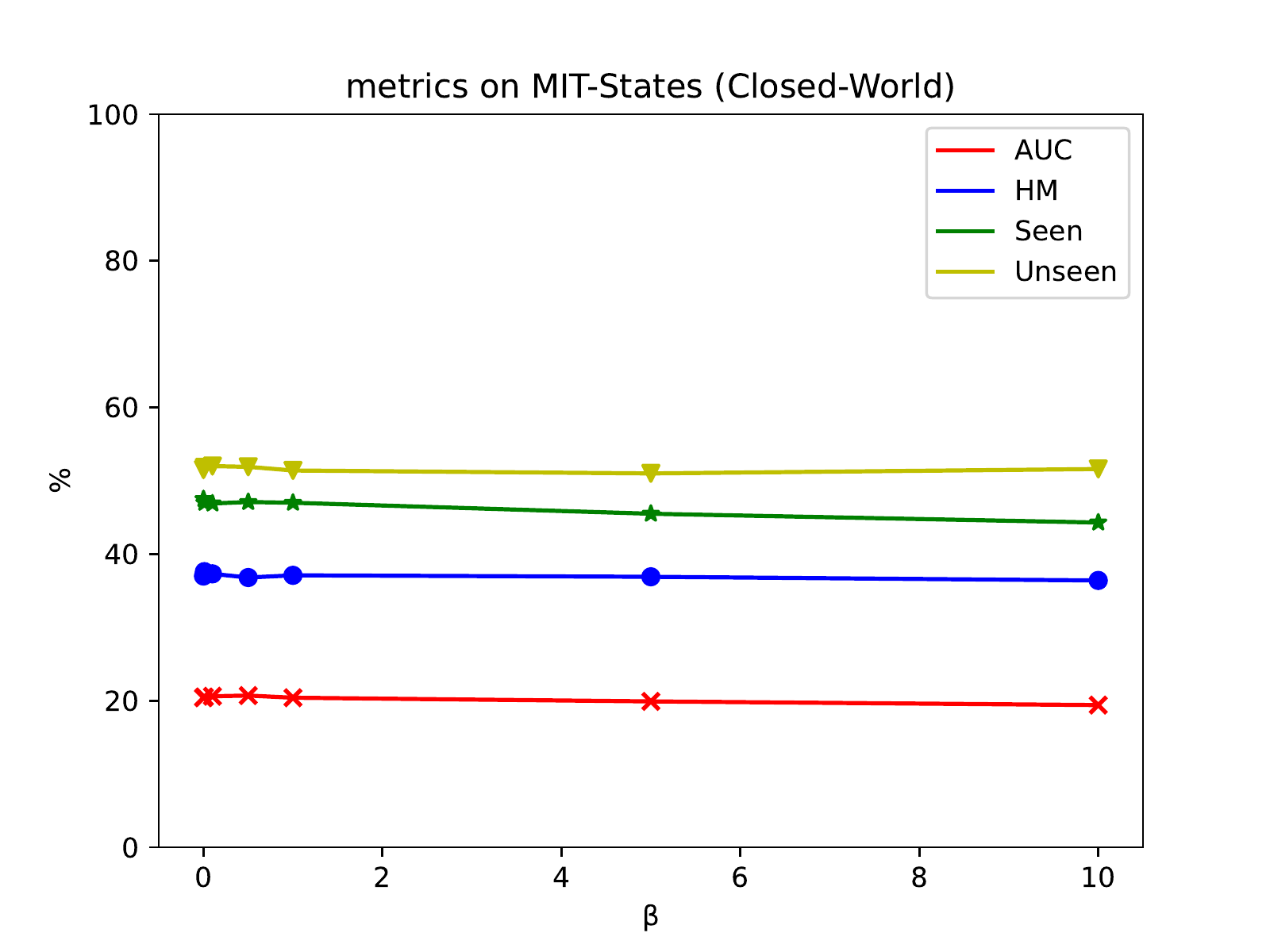}
		\caption{Closed-World ($\beta$)}
		\label{me_ut_b_cw}
	\end{subfigure}
    \begin{subfigure}{0.24\linewidth}
		\centering
		\includegraphics[width=0.95\linewidth]{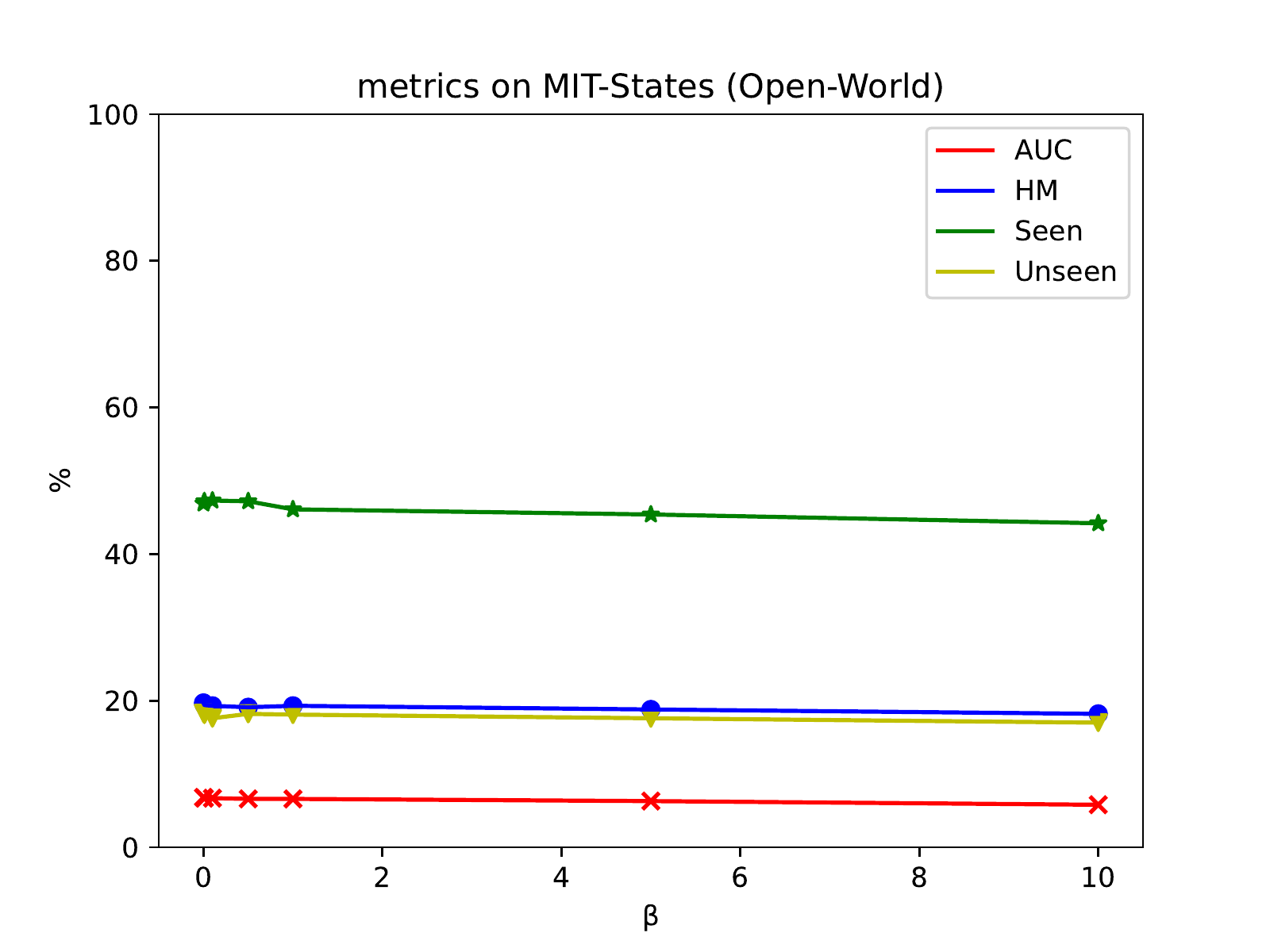}
		\caption{Open-World ($\beta$)}
		\label{me_ut_b_ow}
	\end{subfigure}
	\caption{Metrics on MIT-States with the settings of Closed-World and Open-World.}
	\label{fig:me_mit}
\end{figure}

The loss function of DFSP is $\mathcal{L} =\mathcal{L}_{dfm} + \alpha \mathcal{L}_{st+obj} + \beta \mathcal{L}_{spm}$, which contains two hyper-parameters $\alpha$ and $\beta$.
$\mathcal{L}_{dfm}$ is the final pair loss in DFM, $\mathcal{L}_{st+obj}$ is the decomposed features pair loss and $\mathcal{L}_{spm}$ is the pair loss before fusion in SPM.
To evaluate the influence of them for DFSP(\textit{t2i}), we show the hyper-parameters analysis in this section on MIT-States and UT-Zappos with the settings of Closed-World and Open-World.
$\alpha$ and $\beta$ are set to seven different orders of magnitude parameters, specifically $\left [ 0,0.01,0.1,0.5,1.0,5.0,10.0 \right ] $.
The results are shown in Fig.~\ref{fig:me_ut} and Fig.~\ref{fig:me_mit}.
Since UT-Zappos consists of less \textit{states} and \textit{objects} in Tab.~\ref{tab:sum_data}, the metrics consist of \textit{S}, \textit{U}, \textit{H} and \textit{AUC} are more sensitive to $\alpha$ and $\beta$, while metrics on MIT-States are going to be smoother.
It can be concluded from the graph that the optimal selection range of $\alpha$ and $\beta$ is $0\sim 1$.

\section{Pseudocode}
We illustrate the core pseudocode in this section, including the decompose and recompose function.
Given the pair\_idx, we can obtain the att\_idx and obj\_idx, which could be utilized to decompose the state feature text\_att and object feature text\_obj.
Additionally, the recompose function is reverse to decompose function and the code is shown as Code 1.

\newpage

\begin{lstlisting}[language=Python,title={Code 1: Decompose and Recompose}]
def decompose(text_feature, pair_idx):
    t, l, c = text_feature.shape
    att_idx, obj_idx = pair_idx[:, 0].cpu().numpy(), pair_idx[:, 1].cpu().numpy()
    text_att = torch.zeros(t, self.attributes, c).cuda()
    text_obj = torch.zeros(t, self.classes, c).cuda()
    for i in range(self.attributes):
        text_att[:, i, :] = text_feature[:, np.where(att_idx==i)[0], :].mean(-2)
    for i in range(self.classes):
        text_obj[:, i, :] = text_feature[:, np.where(obj_idx==i)[0], :].mean(-2)    
    text_feature_plus = torch.cat([text_att, text_obj], dim=1)
    return text_feature_plus


def recompose(text_feature_plus, pair_idx):
    t, l, c = text_feature.shape
    att_idx, obj_idx = pair_idx[:, 0].cpu().numpy(), pair_idx[:, 1].cpu().numpy()
    text_com_feature = torch.zeros(t, len(idx), c).cuda()
    text_com_feature = text_feature[:, att_idx, :] * text_feature[:, obj_idx + offset, :]
    return text_feature
\end{lstlisting}

\section{Qualitative Results}
\begin{figure}
    \centering
    \includegraphics{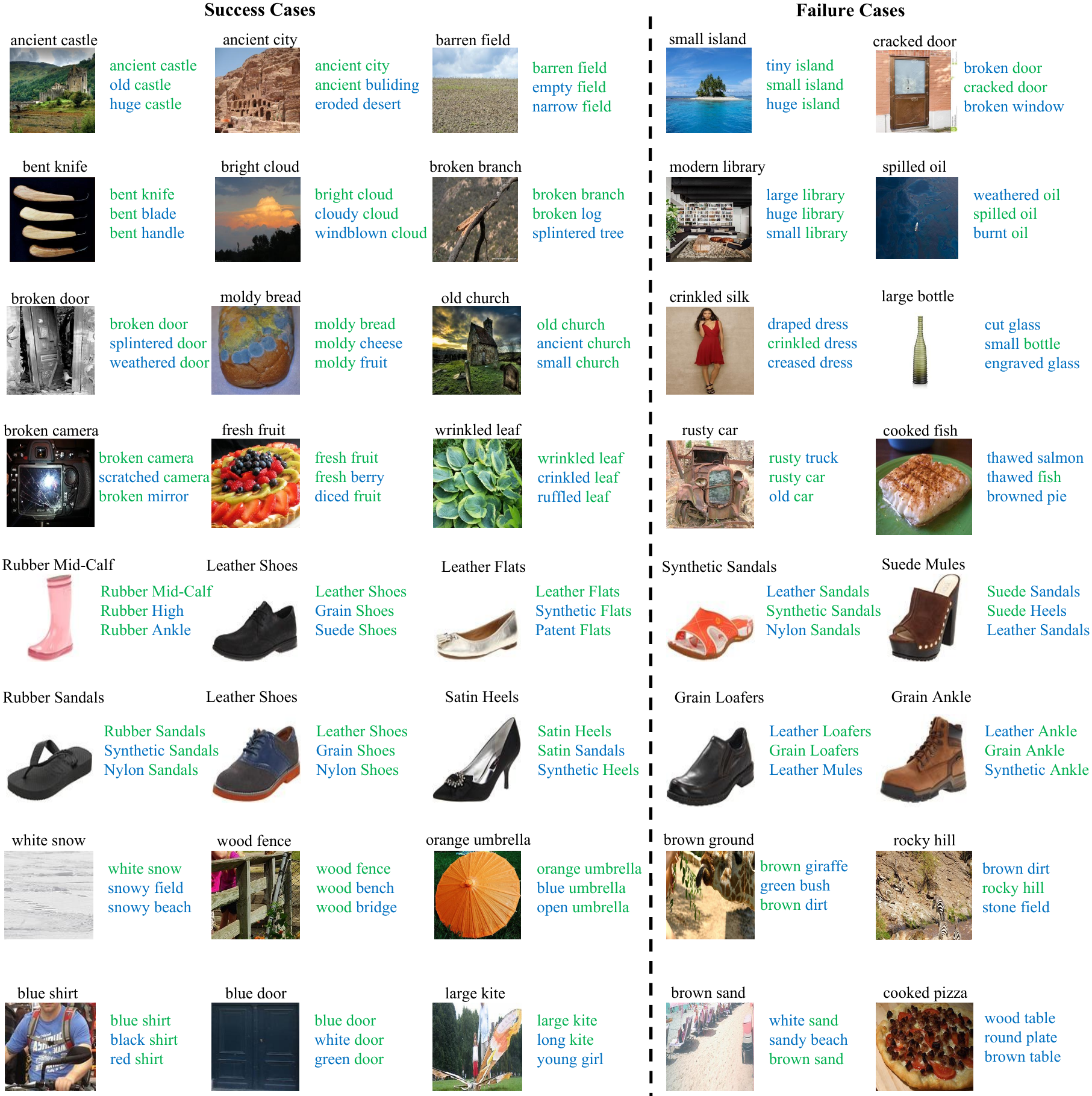}
    \caption{Qualitative top-3 results on MIT -States~\cite{isola2015discovering}, UT-Zappos~\cite{yu2014fine} and C-GQA~\cite{naeem2021learning}. \textcolor{blue}{blue} denotes the wrong prediction and \textcolor{green}{green} represents the right case. The three columns on the left are success cases, and the two columns on the right are wrong cases for the top-1 prediction.}
    \label{fig:top3}
\end{figure}

We report the top-1 qualitative results in the body of the paper.
To better prove the effectiveness of DFSP, we show the top-3 qualitative results on MIT-States, UT-Zappos and CGQA in this section, which can be seen in Fig.~\ref{fig:top3}.
From the prediction results of top-3, it can be seen that even if top-1 has no successful cases, most of top-3 results can be predicted correctly.
The compositions that the model has not seen can still be predicted correctly, which proves the generalization ability of the model to unseen concepts.
Due to the abstract nature of state, it is more difficult to identify a state than an object, which can be seen more failure cases in Fig.~\ref{fig:top3} for states.

{\small
\bibliographystyle{ieee_fullname}
\bibliography{egbib}
}